\documentclass{article} 
\usepackage{iclr2027_conference,times}

\usepackage{amsmath,amsfonts,bm}

\def\eqref#1{equation~\ref{#1}}

\def\1{\bm{1}}

\DeclareMathAlphabet{\mathsfit}{\encodingdefault}{\sfdefault}{m}{sl}
\SetMathAlphabet{\mathsfit}{bold}{\encodingdefault}{\sfdefault}{bx}{n}

\usepackage{hyperref}
\usepackage{url}

\usepackage{microtype}      
\usepackage{xcolor}         
\usepackage{amsmath}
\usepackage{booktabs}
\usepackage{multirow}
\usepackage{adjustbox}
\usepackage{graphicx}
\usepackage{subcaption}
\usepackage[table]{xcolor}
\usepackage{array}

\title{From Experts to Sub-experts: Fine-grained Parameter-Efficient Fine-Tuning for MoE LLMs}

\author{
Zhentao Tan$^{1}$\thanks{These authors contributed equally to this work.} \quad
Chang Liu$^{1,2}$\footnotemark[1] \quad
Yao Liu$^{1}$ \quad
Yue Wu$^{1}$ \quad
Jieping Ye$^{1}$ \\
$^{1}$Alibaba Group \\
$^{2}$Southeast University
}

\iclrfinalcopy 
\begin{document}

\maketitle
\fancyhead{}                         
\renewcommand{\headrulewidth}{0pt}  

\begin{abstract}
As large language models (LLMs) scale rapidly, dense full-parameter adaptation becomes increasingly expensive, motivating sparse and modular architectures such as Mixture-of-Experts (MoE) models. This shift raises a key question for parameter-efficient fine-tuning (PEFT): at what granularity should parameters be selected and updated? Existing PEFT methods such as LoRA operate on predefined weight matrices, while expert-level sparse tuning methods update entire selected experts. However, we observe that activated experts are internally sparse, with only a small fraction of intermediate channels strongly responding to downstream tasks, indicating that expert-level adaptation is still too coarse. We propose \textbf{NSFT} (\emph{Neural Sub-expert Fine-Tuning}), a fine-grained PEFT framework that refines MoE adaptation from experts to sub-experts. NSFT decomposes each expert along the intermediate dimension into structured channel groups and selects task-relevant sub-experts by combining routing importance with intra-expert activation saliency. To optimize sparse partial updates, NSFT further introduces learning-rate scaling and dynamic gradient scaling to compensate for the reduced effective update magnitude. Experiments on OLMoE and Ling-mini-2.0 across challenging domain-specific tasks and general benchmarks show that NSFT consistently outperforms representative PEFT and expert-level sparse tuning baselines, while using substantially fewer trainable parameters and preserving competitive general capability. These results suggest that sub-expert-level adaptation is a more precise and efficient PEFT paradigm for MoE LLMs. Our code is available at \url{https://github.com/aheadformore/NSFT}.
\end{abstract}

\section{Introduction}

As large language models (LLMs) scale into the hundreds of billions---and increasingly, trillions---of parameters, the conventional paradigm of dense, full-parameter training faces mounting computational and memory bottlenecks \citep{bai2023qwen,grattafiori2024llama}. This pressure has catalyzed a fundamental shift from dense parameterization toward sparse, modular architectures \citep{jiang2024mixtral,yang2025qwen3,liu2024deepseek,team2025kimi}. Along with this architectural transition, parameter-efficient adaptation raises a crucial question: \emph{at what granularity should model parameters be selected and updated for effective downstream learning?} The choice of adaptation granularity not only affects parameter efficiency and training cost, but also relates to broader questions about modularity in Transformers \citep{geva2021transformer}, sparsity in knowledge representation \citep{hu2022lora,liu2024dora}, and the intrinsic redundancy of model parameters \citep{frankle2018lottery,cheng2025mixture}.

\begin{figure*}[t!]
  \centering
  \includegraphics[width=\linewidth]{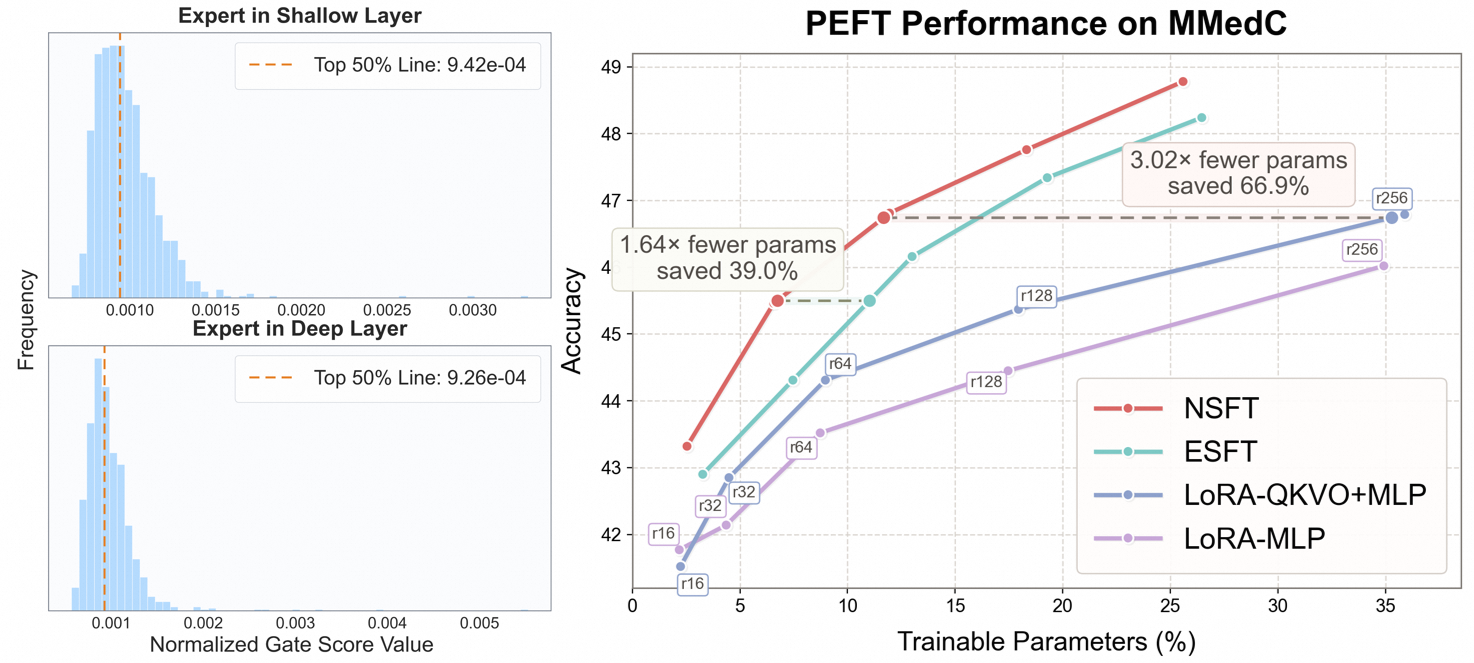}
  \caption{Left: channel-level gate-score distributions of shallow and deep OLMoE experts on MMedC, showing strong intra-expert sparsity. The dashed line marks the top-50\% cumulative activation threshold. Right: PEFT performance on MMedC under different trainable-parameter budgets, where NSFT achieves a better accuracy--parameter trade-off than ESFT and LoRA variants.}
  \label{subexperts_activation}
  \vspace{-0.2cm}
\end{figure*}

Historically, adaptation granularity has largely followed architectural topology. In dense LLMs, parameter-efficient fine-tuning (PEFT) methods such as LoRA~\citep{hu2022lora} typically operate on linear weight matrices, approximating task-specific updates with low-rank components. However, such matrix-level adaptation may spread gradients across broadly coupled neurons, making it difficult to precisely target the functional components most relevant to downstream tasks. Mixture-of-Experts (MoE) architectures introduce routed computation paths~\citep{fedus2022switch}, enabling a coarser form of sparsity at the expert level. Recent methods such as ESFT~\citep{wang2024let} therefore update only selected experts for modular adaptation. Nevertheless, expert-level selection remains too coarse. As shown in Figure~\ref{subexperts_activation} (left), channel activations inside activated experts are highly concentrated near zero, indicating that only a small fraction of internal channels are strongly involved in the target task (refer to the Appendix~\ref{intra_expert_appendix} for more details). Updating the whole expert therefore introduces substantial redundancy and limits the parameter efficiency of expert-level tuning.

\begin{figure*}[t]
  \centering
  \includegraphics[width=\linewidth]{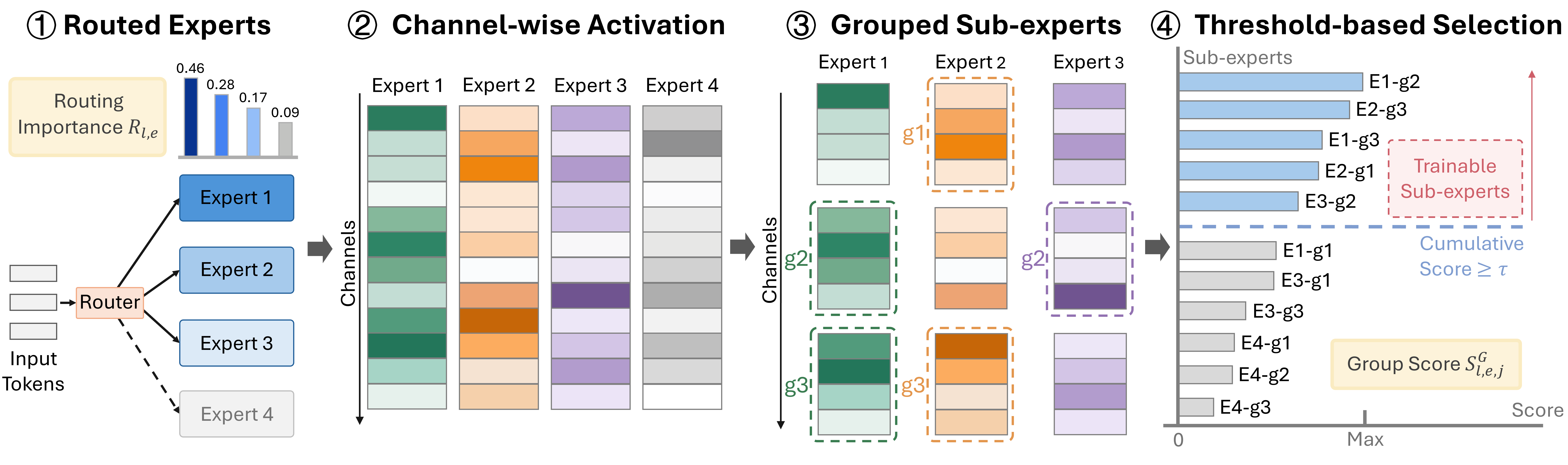}
  \caption{Illustration of sub-expert selection. Routed experts are first identified, and channel-wise activations are computed within each expert. With group size $G=4$, consecutive channels are partitioned into sub-expert groups, where $E1$-$g1$ denotes the first group of Expert 1. All groups are then scored and ranked, and the groups selected under the threshold $\tau$ are used as trainable sub-experts for fine-tuning.}
  \label{lct}
  \vspace{-0.2cm}
\end{figure*}

To address this limitation, we propose \textbf{NSFT} (\emph{Neural Sub-expert Fine-Tuning}), a fine-grained PEFT method that moves sparse adaptation below the expert level. As illustrated in Figure~\ref{lct}, NSFT decomposes each expert along its intermediate dimension into structured sub-experts and selects the most task-relevant ones for fine-tuning. Inspired by MoNE~\citep{cheng2025mixture}, our method treats sub-experts as compact functional units rather than updating entire experts. By focusing updates on responsive internal channel groups, NSFT constructs sparse and precise adaptation pathways while avoiding redundant updates to weakly activated expert parameters.

However, fine-grained sub-expert tuning also introduces a new optimization challenge. Since only a small subset of channels inside each expert is updated, the effective update magnitude becomes much smaller than that of full-expert or expert-level tuning. If the original optimization settings are used directly, the selected sub-experts may be under-updated, limiting adaptation effectiveness. To address this issue, we further introduce an adaptive training strategy with two complementary components: learning-rate scaling, which compensates for the reduced global update step under sparse partial updates, and gradient scaling, which redistributes update strength among selected sub-experts according to their relative importance.

We evaluate NSFT on two MoE backbones, \textbf{OLMoE}~\citep{muennighoff2024olmoe} and \textbf{Ling-mini-2.0}~\citep{inclusionai2025lingmini20}, across a wide range of challenging domain-specific adaptation tasks, including multilingual medicine~\citep{qiu2024towards}, scientific information extraction~\citep{wadden2025sciriff}, domain-specific RAG~\citep{liu2025rag}, mathematical reasoning~\citep{tian2025correctanswersequaldistillation}, code generation~\citep{olmo2025olmo}, and table question answering~\citep{wu2025tablebench}. These tasks cover diverse cognitive demands, such as factual recall, logical reasoning, instruction following, and structured output synthesis, providing a comprehensive testbed for evaluating parameter-efficient adaptation. 

Our results show that conventional PEFT and sparse tuning methods still face clear efficiency--performance limitations. As shown in Figure~\ref{subexperts_activation} (right), NSFT achieves a better accuracy--parameter trade-off on MMedC than ESFT and LoRA variants, reaching higher accuracy with substantially fewer trainable parameters. In particular, NSFT saves up to \textbf{66.9\%} trainable parameters compared with LoRA-QKVO+MLP under comparable or stronger performance, and requires fewer parameters than ESFT to reach the same accuracy level. Across OLMoE and Ling-mini, NSFT consistently improves domain adaptation over ESFT and LoRA while preserving competitive general capability, demonstrating that fine-grained sub-expert selection provides a more effective use of the trainable parameter budget.

Our contributions are threefold:
\begin{itemize}
    \item Conceptual: We revisit the adaptation granularity of MoE models and show that expert-level sparse tuning remains too coarse for domain-specific fine-tuning. By shifting the update unit from entire experts to sub-experts, we provide a finer-grained perspective on how task-relevant knowledge is localized and adapted within MoE architectures.
    \item Methodological: We design and implement the Sub-experts framework, enabling fine-grained, structured sparsity through principled decomposition of expert modules, thereby advancing the frontier of efficient adaptation in MoE models.
    \item Empirical: We demonstrate across diverse real-world domains and two MoE backbones that NSFT consistently outperforms representative PEFT and expert-level sparse tuning baselines, while requiring substantially fewer trainable parameters. These results highlight the effectiveness of fine-grained sub-expert adaptation as a more parameter-efficient and performance-competitive approach for customizing large-scale MoE LLMs.
\end{itemize}

\section{Related Works}

\textbf{Adaptation Granularity in Dense LLMs.}
The pursuit of efficient LLM adaptation has shifted attention from full-parameter updates to the choice of trainable granularity within model architectures. In dense models, existing methods mainly define update units from three perspectives. First, \textbf{externally added modules}, such as Adapters~\citep{houlsby2019parameter,wang2022adamix} and Soft Prompting~\citep{li2021prefix,zhang2023towards}, introduce learnable auxiliary components while keeping the backbone frozen. Second, reparameterization-based methods operate on \textbf{individual linear layers}; LoRA~\citep{hu2022lora}, DoRA~\citep{liu2024dora}, and their successors~\citep{wu2025s,che2025lora} approximate weight updates with low-rank decompositions. Third, selective fine-tuning methods directly update subsets of existing parameters~\citep{xu2021raise,hui2025hft,he2023sensitivity}. Despite their differences, most dense-model PEFT methods remain closely tied to linear transformations, treating Transformers more as collections of tunable operators than as modular functional units.

\textbf{Expert-Level Adaptation in Sparse Architectures.}
MoE architectures introduce structural sparsity through routing and expert modules, naturally enabling coarser-grained adaptation at the expert level~\citep{liu2024deepseek,team2025kimi,dai2024deepseekmoe}. In this setting, the \textbf{expert} becomes an intuitive update unit. Expert-Specialized Fine-Tuning (ESFT)~\citep{wang2024let} updates only experts that strongly respond to target data. DES-MoE~\citep{li2025dynamic} further adjusts expert selection through a multi-stage process for multi-domain adaptation, while DR-LoRA~\citep{deng2026dr} uses non-uniform expert activation to assign different LoRA ranks within experts. These studies show that expert-level adaptation can improve efficiency for sparse MoE models, but they still treat each selected expert as an indivisible unit.

\textbf{From Full Experts to Sub-Expert Units.}
Recent evidence suggests that experts themselves remain internally redundant. Mixture of Neuron Experts (MoNE)~\citep{cheng2025mixture} shows that experts can be decomposed into neuron-level grains, where activating only high-response neurons can preserve inference performance. However, directly tuning at the neuron level can be overly fragmented, leading to unstable optimization and weak update capacity. We therefore introduce \emph{sub-experts} as fine-grained but structured update units. By grouping neurons into cohesive sub-expert units, our method retains the flexibility of neuron-level selection while avoiding the instability of isolated-neuron updates, providing a more robust paradigm for parameter-efficient adaptation of MoE LLMs.

\vspace{-0.1cm}
\section{Method}
\vspace{-0.1cm}
\label{sec: method}
\subsection{Preliminaries}
\vspace{-0.1cm}
\paragraph{Mixture-of-Experts Architecture.}
We first revisit the standard MoE architecture. Let $x\in\mathbb{R}^{d_{\mathrm{model}}}$ denote the input hidden state to layer $l$. An MoE layer contains $N_E$ experts $\{E_{l,e}\}_{e=1}^{N_E}$ and a router that assigns routing weights to experts. For a GLU-style expert, we write
\begin{equation}
    E_{l,e}(x)
    =
    \mathbf{W}_{\mathrm{down}}^{l,e}
    \left(
    \mathrm{SiLU}\!\left(g_{l,e}(x)\right)
    \odot
    \mathbf{W}_{\mathrm{up}}^{l,e}x
    \right),
\end{equation}
where $g_{l,e}(x)
    =
    \mathbf{W}_{\mathrm{gate}}^{l,e}x$.
Here, $\mathbf{W}_{\mathrm{gate}}^{l,e}, \mathbf{W}_{\mathrm{up}}^{l,e}\in\mathbb{R}^{d_{\mathrm{expert}}\times d_{\mathrm{model}}}$ and $\mathbf{W}_{\mathrm{down}}^{l,e}\in\mathbb{R}^{d_{\mathrm{model}}\times d_{\mathrm{expert}}}$.

The router produces expert weights $\mathbf{p}_l(x)\in\mathbb{R}^{N_E}$ and selects the top-$K$ experts. Let $\mathcal{E}_l(x)$ denote the selected expert set. The MoE output is then 
\begin{equation}
    \mathrm{MoE}_l(x)
    =
    \sum_{e\in \mathcal{E}_l(x)}
    p_{l,e}(x) E_{l,e}(x).
\end{equation}

\vspace{-0.2cm}
\paragraph{Neuron Expert Decomposition.}
Recent studies such as MoNE~\citep{cheng2025mixture} show that each MoE expert can be decomposed into finer-grained \emph{neuron experts}. For a GLU-style expert, the output of expert $e$ in layer $l$ can be expressed as $E_{l,e}(x)
    =
    \sum_{m=1}^{d_{\mathrm{expert}}}
    a_{l,e,m}(x)
    \mathbf{A}_{l,e,m}x$,
where $a_{l,e,m}(x)
    =
    \mathrm{SiLU}\!\left(g_{l,e}(x)\right)_m$
is the activation of the $m$-th intermediate channel, and
    $\mathbf{A}_{l,e,m}
    =
    \mathbf{W}_{\mathrm{down}}^{l,e}[:,m]
    \mathbf{W}_{\mathrm{up}}^{l,e}[m,:]$
is the rank-one transformation induced by this channel. This formulation reveals that an expert is a superposition of neuron-wise rank-one transformations, motivating us to select and update task-relevant neuron groups as sub-experts rather than treating the entire expert as an indivisible unit.

\vspace{-0.1cm}
\subsection{Fine-grained Sub-expert Selection}
\label{sub-experts-selection}
\vspace{-0.1cm}
Existing expert-level sparse adaptation methods, such as ESFT, improve parameter efficiency by updating only a small subset of experts during supervised fine-tuning. However, they still treat each selected expert as an indivisible unit. In practice, the parameter space inside an activated expert remains highly sparse: only a small portion of its intermediate channels are consistently responsive to the downstream task, while many others contribute marginally. As a result, even after expert selection, updating the full expert may still introduce substantial redundancy.

This motivates us to refine the granularity of sparse adaptation from the \emph{expert level} to the \emph{sub-expert level}. Instead of treating each expert as a monolithic MLP block, we decompose it along the intermediate dimension into multiple finer-grained sub-experts. Such a decomposition is natural for MoE feed-forward blocks, where each expert is implemented as a gated MLP and its intermediate channels already exhibit highly non-uniform activation patterns. As shown in Figure~\ref{lct}, we organize consecutive intermediate channels into fixed-size groups, where each group contains $G$ channels and serves as a candidate sub-expert. This grouping provides a structured adaptation unit that is finer than a full expert but less fragmented than individual neurons. Compared with expert-level selection, this finer-grained formulation further removes weakly activated channels inside selected experts and yields a more faithful approximation to the effective task-specific computation path.

Formally, for layer $l$ and expert $e$, let $\mathcal{T}_{l,e}$ denote the set of tokens routed to this expert, and let $m \in \{1,\dots,d_{\mathrm{expert}}\}$ index the intermediate channels of the expert. We first accumulate the channel-wise gate responses over all routed tokens:
\begin{equation}
    M_{l,e,m}
    =
    \sum_{x\in\mathcal{T}_{l,e}}
    \left|
    \mathrm{SiLU}\!\left(g_{l,e}(x)\right)_m
    \right|.
\end{equation}
To characterize the relative importance distribution inside each expert, we normalize the accumulated responses as $ \hat{M}_{l,e,m}=\frac{M_{l,e,m}}{\sum_{m'} M_{l,e,m'}}$. In addition, let $N_{l,e}$ denote the number of times expert $e$ is selected by the router in layer $l$. We define the expert-level routing importance as $R_{l,e}=\frac{N_{l,e}}{\sum_{e'} N_{l,e'}}$.

Based on both expert-level activity and intra-expert channel saliency, we define the channel-level sub-expert importance score as:
\begin{equation}
    S_{l,e,m}=R_{l,e}\hat{M}_{l,e,m}.
\end{equation}

To obtain structured sub-experts, we further aggregate channel scores within each group. Let $\mathcal{G}_{l,e,j}$ denote the $j$-th group of $G$ consecutive channels in expert $e$ of layer $l$. Its group-level importance score is computed as $S_{l,e,j}^{G}
=\sum_{m \in \mathcal{G}_{l,e,j}} S_{l,e,m}$. When $G=1$, the selection reduces to individual channel selection; larger $G$ values produce more structured but coarser sub-expert units.

This score jointly captures whether an expert is globally important for the current task and whether a specific sub-expert group is locally important within that expert. After obtaining $S_{l,e,j}^{G}$, we perform layer-wise selection by flattening all sub-expert groups in the same layer and ranking them in descending order. For layer $l$, we select the smallest subset whose cumulative score exceeds a threshold $\tau$: $\sum_{i=1}^{K_l} \left(S_l^{G}\right)^{(i)} \ge \tau$, where $\{(S_l^{G})^{(i)}\}$ denotes the sorted group-level sub-expert scores in layer $l$. The selected subset is then used as the sub-experts for downstream tuning. In this way, our method preserves the most task-relevant internal channel groups of activated experts while filtering out low-contribution parameters.

\subsection{Adaptive Training for Sub-expert Fine-tuning}

After identifying the selected sub-experts via Section~\ref{sub-experts-selection},
fine-tuning only a small subset of channels within each expert creates
two optimization issues. For expert $e$ in layer $l$, let $\mathcal{A}_{l,e}$
denote the set of selected intermediate channels induced by the selected
sub-expert groups, and let $K_{l,e}=|\mathcal{A}_{l,e}|$ be the number of
selected channels. First, since only $K_{l,e}$ out of $d_{\mathrm{expert}}$ channels
are updated, the effective update magnitude becomes much smaller than in
full-expert tuning, leading to \emph{step-size underestimation}.
Second, even among the selected channels, their task relevance is
heterogeneous, while standard optimization would treat them uniformly.
We address these two issues with learning-rate scaling and gradient
scaling, respectively.

\vspace{-0.2cm}
\paragraph{Learning-rate scaling.}
We first compensate for the loss of global update magnitude caused by
partial updates. We scale the base learning rate according
to the inverse active ratio:
\begin{equation}
    \eta_{l,e}
    =
    \eta_{\mathrm{base}}
    \times
    \min\!\left(\frac{d_{\mathrm{expert}}}{K_{l,e}},\ \alpha_{\max}\right),
    \label{eq:lr_scale_final}
\end{equation}
where $\alpha_{\max}$ is a clipping threshold for stability. Intuitively,
when only a fraction $K_{l,e}/d_{\mathrm{expert}}$ of channels is trainable, each selected
channel should receive a proportionally larger update so that the overall
optimization progress remains comparable to full-expert tuning.

\vspace{-0.2cm}
\paragraph{Gradient scaling.}
Learning-rate scaling restores the \emph{global} update magnitude, but
it does not distinguish the relative importance of different selected
channels. We therefore further modulate the gradient inside each selected
expert by applying a channel-group-level scaling mask during the backward
pass. Let $\mathcal{G}_{l,e}$ denote the set of selected channel groups.
For each parameter tensor, we define a mask $\mathbf{m}_{l,e}$ whose
entries are zero on frozen channels and equal to a group-specific scaling
coefficient $\sigma_g$ on channels belonging to group
$g \in \mathcal{G}_{l,e}$. The resulting masked gradient is $\tilde{\mathbf{g}}_\theta = \nabla_\theta \mathcal{L} \odot \mathbf{m}_{l,e}$.

The scaling coefficient $\sigma_g$ is determined by the relative importance
of each selected group. We first compute the captured importance mass of the
selected channels: $\rho_{l,e}
    =
    \sum_{m \in \mathcal{A}_{l,e}}
    \hat{M}_{l,e,m}$.
For each selected group $g$, we define its activation energy as $g^{(g)}_{\mathrm{energy}}=
    \sum_{m \in g}
    \hat{M}_{l,e,m}$,
and compute the mean group energy: $\bar{g}_{\mathrm{energy}}
    =
    \rho_{l,e}/|\mathcal{G}_{l,e}|$. The group-specific scaling coefficient is then given by
\begin{equation}
    \sigma_g
    =
    \mathrm{clip}\!\left(
        \left(
        \frac{g^{(g)}_{\mathrm{energy}}}{\bar{g}_{\mathrm{energy}}}
        \right)^{\gamma},
        \ 1.0,\ \sigma_{\max}
    \right).
    \label{eq:sigma_final}
\end{equation}
This design amplifies gradients for groups whose activation energy is higher
than the average selected group, while avoiding excessive scaling through
clipping. In this way, gradient scaling redistributes update strength among
selected sub-experts according to their relative task relevance.

\paragraph{Entropy-adaptive modulation.}
To avoid overly sharp or overly flat gradient allocation, we further
control the contrast among groups with an entropy-adaptive exponent:
\begin{equation}
    \gamma
    =
    0.5 + 0.5 \cdot \frac{H(\mathcal{G}_{l,e})}{H_{\max}},
    \quad
    H(\mathcal{G}_{l,e})
    =
    -\sum_g \hat{p}_g \log \hat{p}_g,
\end{equation}
where
\begin{equation}
    \hat{p}_g
    =
    \frac{g^{(g)}_{\mathrm{energy}}}{\rho_{l,e}},
    \quad
    H_{\max}
    =
    \log |\mathcal{G}_{l,e}|.
\end{equation}
When the group energy distribution is uniform, $\gamma$ approaches $1$,
preserving inter-group contrast. When it is highly concentrated,
$\gamma$ moves closer to $0.5$, reducing extreme disparities and
improving training stability.

\paragraph{Static and dynamic gradient scaling.}
A straightforward implementation is to compute the scaling coefficients once
from pre-collected importance statistics and keep them fixed throughout
fine-tuning. However, such static scaling can be partially absorbed by Adam-style
adaptive optimization. Adam maintains the first-order moment $m_t$ and
second-order moment $v_t$ of gradients, and updates parameters using the
normalized direction $m_t/\sqrt{v_t}$. Therefore, when gradients are multiplied
by a fixed coefficient, both $m_t$ and $v_t$ are scaled accordingly, making the
normalized update relatively insensitive to this constant factor. As a result,
the intended effect of static gradient scaling may be weakened during training.

To address this issue, we further introduce dynamic gradient scaling. Instead
of using a fixed scaling mask, we update channel activation statistics online
and periodically refresh the scaling coefficients with EMA smoothing. This
allows the gradient allocation to track the evolving importance of selected
sub-experts while avoiding overly noisy updates. In this way, dynamic scaling
acts as an adaptive refinement of static scaling rather than a separate
equivalent variant.

Overall, learning-rate scaling and gradient scaling play complementary roles:
the former restores the global step size under sparse updates, while the latter
allocates gradient strength among selected channels according to their relative
task relevance, with dynamic scaling further improving this allocation during
training.


\definecolor{traincol}{RGB}{245,245,245}

\begin{table*}[t]
\centering
\scriptsize
\setlength{\tabcolsep}{1.pt}
\renewcommand{\arraystretch}{1.12}
\begin{adjustbox}{width=\textwidth}
\begin{tabular}{lccccccccccccccc}
\toprule
\multirow{2}{*}{\textbf{Method}}
& \multicolumn{2}{c}{\textbf{MMedC}}
& \multicolumn{3}{c}{\textbf{SciRiFF}}
& \multicolumn{2}{c}{\textbf{RAG-Instruct}}
& \multicolumn{2}{c}{\textbf{MATH}}
& \multicolumn{2}{c}{\textbf{Code}}
& \multicolumn{2}{c}{\textbf{TableQA}}
& \multirow{2}{*}{\textbf{Avg.}}
& \multirow{2}{*}{\textbf{\shortstack{General \\ Avg.}}} \\
\cmidrule(lr){2-3}
\cmidrule(lr){4-6}
\cmidrule(lr){7-8}
\cmidrule(lr){9-10}
\cmidrule(lr){11-12}
\cmidrule(lr){13-14}
& \cellcolor{traincol} Train. & Domain
& \cellcolor{traincol} Train. & 4k & 8k
& \cellcolor{traincol} Train. & Domain
& \cellcolor{traincol} Train. & Domain
& \cellcolor{traincol} Train. & Domain
& \cellcolor{traincol} Train. & Domain
& & \\
\midrule
Base-OLMoE & \cellcolor{traincol} - & 35.09 & \cellcolor{traincol} - & 27.67 & 25.87 & \cellcolor{traincol} - & 63.11 & \cellcolor{traincol} - & 56.74 & \cellcolor{traincol} - & 39.19 & \cellcolor{traincol} - & 7.08 & 36.39 & 47.49 \\
Full FT & \cellcolor{traincol} 100\% & 52.12 & \cellcolor{traincol} 100\% & 47.14 & 41.95 & \cellcolor{traincol} 100\% & 72.92 & \cellcolor{traincol} 100\% & 59.76 & \cellcolor{traincol} 100\% & 44.77 & \cellcolor{traincol} 100\% & 30.70 & 49.91 & 35.98 \\

\specialrule{0.4pt}{1.5pt}{2pt}

LoRA-r32 & \cellcolor{traincol} 4.49\% & 42.84 & \cellcolor{traincol} 4.49\% & 35.60 & 35.36 & \cellcolor{traincol} 4.49\% & 64.56 & \cellcolor{traincol} 4.49\% & 55.68 & \cellcolor{traincol} 4.49\% & 40.67 & \cellcolor{traincol} 4.49\% & 15.15 & 41.41 & 44.13 \\
LoRA-r64 & \cellcolor{traincol} 8.97\% & 44.31 & \cellcolor{traincol} 8.97\% & 38.56 & 38.07 & \cellcolor{traincol} 8.97\% & 65.62 & \cellcolor{traincol} 8.97\% & 55.70 & \cellcolor{traincol} 8.97\% & 42.13 & \cellcolor{traincol} 8.97\% & 19.26 & 43.38 & 43.41 \\
ESFT-OLMoE & \cellcolor{traincol} 5.64\% & 44.31 & \cellcolor{traincol} 7.46\% & 40.15 & 38.08 & \cellcolor{traincol} 11.37\% & 64.47 & \cellcolor{traincol} 5.00\% & 56.28 & \cellcolor{traincol} 4.00\% & 41.19 & \cellcolor{traincol} 5.55\% & 19.47 & 43.42 & \textbf{46.24} \\
NSFT-OLMoE & \cellcolor{traincol} 4.87\% & \textbf{45.81} & \cellcolor{traincol} 6.57\% & \textbf{44.38} & \textbf{42.05} & \cellcolor{traincol} 10.35\% & \textbf{70.36} & \cellcolor{traincol} 4.29\% & \textbf{57.36} & \cellcolor{traincol} 3.21\% & \textbf{42.24} & \cellcolor{traincol} 4.75\% & \textbf{24.13} & \textbf{46.62} & 45.96 \\
\bottomrule
\end{tabular}
\end{adjustbox}
\caption{In-domain fine-tuning results on OLMoE under threshold $\tau=0.2$. Each domain column reports the result after fine-tuning on the corresponding domain dataset, while \textbf{Train.} denotes the percentage of trainable parameters. Avg. and General Avg. denote the average in-domain performance and the average general-benchmark performance, respectively.}
\label{tab:domain_results_split}
\vspace{-4pt}
\end{table*}

\section{Experiments}
\label{sec: experi}
\vspace{-4pt}

\subsection{Implementation Details}
\label{implementation_experiments}

\paragraph{Fine-tuning datasets.}
We evaluate our method on multiple domain-specific supervised fine-tuning datasets covering five application domains: mathematics, medicine, scientific reasoning, retrieval-augmented question answering, and code generation. Specifically, they include \textbf{MATH}, \textbf{MMedC} \citep{qiu2024towards}, \textbf{SciRIFF} \citep{wadden2025sciriff}, \textbf{RAGQA}, and \textbf{Code}. For RAGQA, we include both the \textbf{PubMedQA} \citep{jin2019pubmedqa} and \textbf{Health} \citep{kotonya2020explainable} subsets. The reported MATH score is the average over \textbf{MATH500} \citep{hendrycks2021measuring} and \textbf{GSM8K} \citep{cobbe2021training}, while the reported Code score is the average over \textbf{HumanEval} \citep{chen2021evaluating} and \textbf{MBPP} \citep{austin2021program}. These datasets are chosen to cover diverse expert activation patterns and task-specific adaptation requirements.

\vspace{-0.2cm}
\paragraph{Evaluation benchmarks.}
To evaluate general capability after fine-tuning, we use four general-purpose benchmarks: \textbf{GPQA}~\citep{rein2023gpqa}, \textbf{MMLU-Redux}~\citep{gema2025we}, \textbf{C-Eval}~\citep{huang2023c}, and \textbf{IFEval}~\citep{zhou2023instruction}. These benchmarks evaluate scientific reasoning, broad knowledge understanding, Chinese knowledge and reasoning, and instruction following, respectively. In all tables, \textbf{General Avg.} denotes the average score over these four benchmarks.

\vspace{-0.2cm}
\paragraph{Baselines.}
We compare our method against three categories of baselines. \textbf{Base}: the original pretrained model without any downstream fine-tuning. \textbf{Full FT}: full fine-tuning of all model parameters, serving as an upper-capacity baseline. \textbf{LoRA-r16} and \textbf{LoRA-r32}: parameter-efficient fine-tuning with LoRA of rank $r=16$ and $r=32$, respectively. \textbf{ESFT}: expert-level sparse fine-tuning, which updates only a subset of experts during adaptation. These baselines allow us to compare our method against dense adaptation, parameter-efficient tuning, and expert-level sparse tuning under a unified setting.

We conduct experiments on two representative MoE language models of different scales: \textbf{OLMoE-7B} \citep{muennighoff2024olmoe} and \textbf{Ling-mini-2.0-16B} \citep{inclusionai2025lingmini20}. Using two models of substantially different sizes allows us to verify that the proposed method is not tied to a specific parameter scale and can generalize across different MoE architectures. We provide more details regarding the experiment in the Appendix~\ref{training_setup_appendix}.

\begin{table*}[t]
\centering
\scriptsize
\setlength{\tabcolsep}{4pt}
\renewcommand{\arraystretch}{1.}
\begin{adjustbox}{width=\textwidth}
\begin{tabular}{llccccccc}
\toprule
\multirow{3}{*}{\textbf{Method}} & \multirow{3}{*}{\textbf{\shortstack{Train. \\ \#Params}}} 
& \multicolumn{4}{c}{\textbf{TableQA Subtasks}} 
& \multirow{3}{*}{\textbf{\shortstack{PoT \\ Avg.}}} 
& \multirow{3}{*}{\textbf{\shortstack{General \\ Avg.}}} \\
\cmidrule(lr){3-6}
& & \shortstack{Instruction\\Type} & \shortstack{Fact\\Checking} & \shortstack{Numerical\\Reasoning} & \shortstack{Data Analysis\\\& Visualization} & & \\
\midrule

Base-Ling  & \cellcolor{traincol} --       & 32.29 & 29.72 &  7.65 & 12.00 & 19.79 & 66.45 \\
Full FT             & \cellcolor{traincol} 100\%    & 48.96 & 53.40 & 26.07 & 30.00 & 39.34 & 62.59 \\
\specialrule{0.4pt}{1.pt}{2pt}
LoRA-r32       & \cellcolor{traincol} 7.45\%   & 39.58 & 40.30 & 16.39 & 20.00 & 28.71 & 64.09 \\
LoRA-r64      & \cellcolor{traincol} 14.84\%  & 43.75 & 43.83 & \textbf{19.64} & 16.00 & 32.00 & 63.90 \\
ESFT-Ling          & \cellcolor{traincol} 3.35\%   & 37.50 & 21.91 &  9.76 & \textbf{24.00} & 17.68 & \textbf{67.03} \\
NSFT-Ling      & \cellcolor{traincol} 2.86\%   & \textbf{48.96} & \textbf{50.88} & 16.34 & 18.00 & \textbf{34.44} & 66.18 \\
\bottomrule
\end{tabular}
\end{adjustbox}
\caption{TableQA results on Ling-mini-2.0. The table reports subtask performance, the average TableQA result (PoT Avg.), and the average general-benchmark performance after fine-tuning.}
\label{tab:tableqa_results}
\end{table*}

\begin{figure*}[t]
  \centering
  \includegraphics[width=\linewidth]{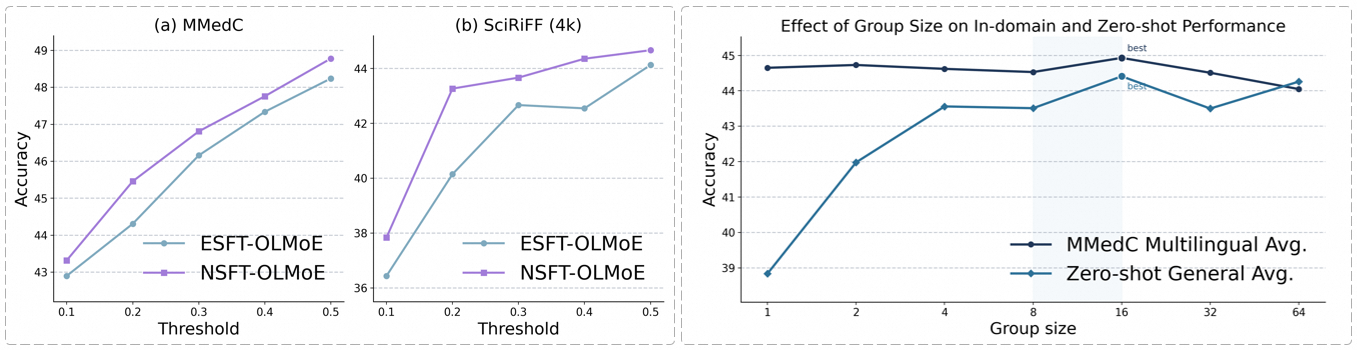}
  \caption{Threshold and group-size ablations. Left: performance under different selection thresholds on MMedC and SciRiFF. The NSFT results correspond to NSFT-SG, which uses learning-rate scaling and static gradient scaling. Right: effect of group size on in-domain (MMedC) and zero-shot general performance.}
  \label{threshold&group}
  \vspace{-0.2cm}
\end{figure*}

\begin{figure*}[t!]
  \centering
  \includegraphics[width=0.95\linewidth]{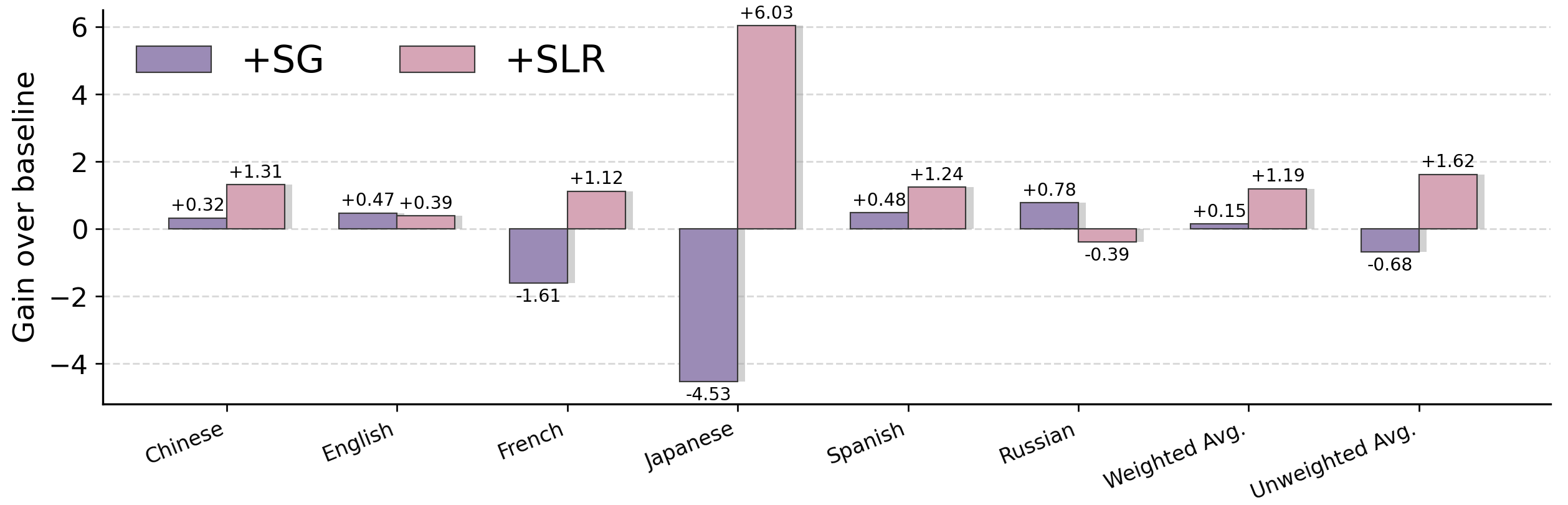}
  \vspace{-0.2cm}
  \caption{Ablation of gradient scaling (+SG) and learning-rate scaling (+SLR) on MMedC multilingual subtasks using OLMoE. Bars show gains over the NSFT baseline, i.e., the variant with only sub-expert selection and no adaptive training strategy.}
  \label{scaling_improvement}
  \vspace{-0.2cm}
\end{figure*}

\subsection{Main Results}
\vspace{-0.1cm}
We evaluate whether NSFT offers a better balance among domain adaptation, parameter efficiency, and general capability preservation than PEFT and expert-level sparse tuning baselines. Results are reported on both OLMoE and Ling-mini across domain-specific and general benchmarks. Unless otherwise specified, NSFT refers to the full method with selection threshold $\tau=0.2$, learning-rate scaling, and dynamic gradient scaling.

\vspace{-0.2cm}
\paragraph{NSFT significantly improves in-domain adaptation on OLMoE while preserving competitive general capability.}
Table~\ref{tab:domain_results_split} summarizes the main results on OLMoE across multiple domain-specific fine-tuning datasets. NSFT-OLMoE achieves the best in-domain performance among efficient adaptation baselines, with an average score of 46.62, substantially outperforming ESFT-OLMoE and both LoRA variants. Full FT obtains a higher in-domain average of 49.91, but its general-benchmark average drops sharply to 35.98. In contrast, NSFT-OLMoE maintains a general average of 45.96, nearly matching ESFT-OLMoE's 46.24, while requiring only \textbf{5.67\%} average trainable parameters. These results show that sub-expert-level adaptation effectively strengthens domain-specific learning without the severe generalization loss caused by full fine-tuning.

\vspace{-0.2cm}
\paragraph{On TableQA, NSFT achieves a markedly better efficiency--performance trade-off than ESFT and LoRA.}
As shown in Table~\ref{tab:tableqa_results}, NSFT-Ling consistently outperforms ESFT-Ling on fine-grained TableQA subtasks. It achieves a PoT average of \textbf{34.44}, substantially higher than ESFT-Ling's 17.68, while using only 2.86\% trainable parameters. Compared with LoRA, NSFT-Ling is also much more parameter-efficient and surpasses the strongest LoRA variant, LoRA-r64. Moreover, it preserves stronger general capability, reaching a general average of 66.18, which is higher than all LoRA variants and clearly above Full FT. These results show that fine-grained sub-expert adaptation provides a better efficiency--performance trade-off for TableQA fine-tuning. For more detailed results and discussions on this part, please refer to the Appendix~\ref{tableqa_appendix}.

These results show that both ESFT and LoRA are suboptimal for domain-specific MoE adaptation. ESFT is too coarse because it updates entire experts, while LoRA applies generic low-rank updates without explicitly targeting task-relevant expert components. In contrast, NSFT selects responsive sub-experts inside activated experts, enabling more precise parameter updates and a better efficiency--generalization trade-off.

\vspace{-0.1cm}
\subsection{Ablation Analysis}
\vspace{-0.1cm}

To investigate the key factors behind NSFT, we conduct ablations on both sub-expert selection and adaptive optimization. We analyze the sensitivity to the selection threshold, the effect of group size $G$, the contribution of learning-rate and gradient scaling, and the difference between static and dynamic scaling. These studies show how fine-grained selection and adaptive training jointly contribute to the effectiveness and stability of NSFT. Additional ablation results, including detailed threshold robustness, group granularity and dynamic scaling analyses, are provided in Appendix~\ref{threshold_appendix}~\ref{group_appendix}~\ref{static_dynamic_appendix}.

\begin{table*}[t]
\centering
\small
\setlength{\tabcolsep}{5pt}
\renewcommand{\arraystretch}{1.15}
\begin{adjustbox}{width=\textwidth}
\begin{tabular}{lccccccc}
\toprule
\multirow{2}{*}{\textbf{Method}}
& \multicolumn{2}{c}{\textbf{MMedC}}
& \multicolumn{3}{c}{\textbf{SciRiFF}}
& \multicolumn{2}{c}{\textbf{RAG-Instruct}} \\
\cmidrule(lr){2-3} \cmidrule(lr){4-6} \cmidrule(lr){7-8}
& \textbf{Avg.} & \textbf{General Avg.}
& \textbf{Avg}$_{4096}$ & \textbf{Avg}$_{8192}$ & \textbf{General Avg.}
& \textbf{Avg.} & \textbf{General Avg.} \\
\midrule
NSFT-SG       & 45.46 & 45.43 & 43.27 & 39.99 & \textbf{45.55} & 68.80 & \textbf{45.47} \\
NSFT-DSG\_iter1      & \textbf{45.89} & \textbf{45.75} & 42.70 & 39.83 & 45.05 & \textbf{70.36} & 44.80 \\
NSFT-DSG\_iter5 & 45.81 & 45.51 & \textbf{44.38} & \textbf{42.05} & 45.46 & \textbf{70.36} & 45.46 \\
\bottomrule
\end{tabular}
\end{adjustbox}
\vspace{-0.2cm}
\caption{Static vs. dynamic gradient scaling using OLMoE. NSFT-SG uses a fixed scaling mask, while NSFT-DSG\_iter1 and NSFT-DSG\_iter5 update the dynamic scaling mask every one and five training steps, respectively. Results are reported on MMedC, SciRiFF, and RAG-Instruct, including in-domain and general-benchmark averages.}
\label{tab:avg_scaling_three_datasets}
\vspace{-0.2cm}
\end{table*}

\vspace{-0.2cm}
\paragraph{Threshold Robustness.}
Figure~\ref{threshold&group} (left) compares ESFT and NSFT under different selection thresholds on MMedC and SciRiFF-4k. NSFT consistently outperforms ESFT across thresholds, showing that sub-expert-level selection is more robust and effective than expert-level sparse tuning under different sparsity budgets. This advantage mainly comes from allocating updates to task-relevant internal channels rather than entire experts. For a fair comparison with ESFT and to balance performance with training cost, we use $\tau=0.2$ for both ESFT and NSFT in the main experiments, following the original ESFT setting.

\vspace{-0.2cm}
\paragraph{Choosing the Sub-expert Group Granularity.}
In our implementation, each expert is partitioned into contiguous channel groups along the intermediate dimension, and each group serves as a candidate sub-expert. Figure~\ref{threshold&group} (right) shows the effect of group size on MMedC multilingual and zero-shot general performance. As $G$ increases from 1 to 16, the zero-shot general average improves steadily, while the in-domain MMedC performance remains stable. The best overall balance is achieved at $G=16$, where the general performance peaks and the in-domain score stays among the highest. This indicates a trade-off between flexibility and expressiveness. Overly small groups fragment the trainable subspace and reduce effective update capacity, while overly large groups introduce redundant channels and weaken fine-grained selection. A moderate group size, especially $G=16$, provides the best balance.

\vspace{-0.2cm}
\paragraph{Effect of Gradient and Learning-rate Scaling.}
Figure~\ref{scaling_improvement} compares the effect of gradient scaling (+SG) and learning-rate scaling (+SLR) on multilingual MMedC tasks. We observe that +SLR consistently improves performance on most languages and yields clear gains on both weighted and unweighted averages, indicating that compensating for the reduced effective step size is critical under partial updates. +SG also improves several languages and the weighted average, although its effect is less stable. This is because, when only a subset of sub-experts is updated, the effective update magnitude becomes significantly smaller; learning-rate scaling directly compensates for this effect at the optimizer level, while gradient scaling further enhances the selected sparse updates. These results confirm the effectiveness of our adaptive scaling strategy.

\vspace{-0.2cm}
\paragraph{Static vs. Dynamic Scaling.}
To evaluate the effect of update frequency, we compare static gradient scaling with dynamic gradient scaling updated every step or every five steps, as shown in Table~\ref{tab:avg_scaling_three_datasets}. Dynamic scaling generally improves in-domain adaptation over static scaling, showing that refreshing the scaling mask during training better captures the evolving importance of selected sub-experts. The best update interval is task-dependent: iter1 performs best on MMedC, while iter5 achieves stronger results on SciRiFF and better general retention on RAG-Instruct. This suggests that dynamic scaling improves over static scaling, but its update frequency should balance rapid adaptation with noise suppression.


\vspace{-0.2cm}
\section{Conclusion}
\vspace{-0.2cm}
We present NSFT, a fine-grained PEFT framework for MoE LLMs that refines sparse adaptation from whole experts to sub-experts. By selecting task-relevant internal channel groups inside activated experts, NSFT enables more precise updates than expert-level sparse tuning. To make such partial updates effective, we further introduce learning-rate scaling and dynamic gradient scaling to compensate for the reduced effective update magnitude. Experiments on OLMoE-7B and Ling-mini-2.0-16B show that NSFT consistently outperforms representative PEFT and expert-level sparse tuning baselines, while using substantially fewer trainable parameters and preserving competitive general capability. These results indicate that expert-level sparsity remains too coarse, and that exploiting fine-grained sub-expert structure is a promising direction for efficient adaptation of MoE LLMs.

\bibliography{iclr2027_conference}
\bibliographystyle{iclr2027_conference}

\appendix
\section{Experimental Setup and Dataset Details}
\label{training_setup_appendix}
\subsection{Training and Evaluation Setup}
We conduct experiments on two Mixture-of-Experts (MoE) models: Ling-mini-2.0 and OLMoE, across seven downstream tasks including TableQA, SciRIFF, Math, Code, RAGQA, MMedC, and Multilingual. All methods are trained for 2 epochs with a global batch size of 8 and a per-device batch size of 1 on 8 GPUs (NVIDIA A100 80G). We use the AdamW optimizer with $\beta_1$ = 0.9, $\beta_2$ = 0.95, and $\epsilon$ = 1e-8. The peak learning rate is set to 3e-5 with a polynomial (linear) decay schedule to a minimum learning rate of 3e-6, preceded by a warmup phase covering 2\% of total training steps. The maximum sequence length is 8,192 tokens, and we employ a bin-packing strategy to concatenate multiple samples into a single sequence for efficient training. All experiments are conducted in BFloat16 precision with FlashAttention-2 and gradient checkpointing enabled. The random seed is fixed at 1,234 for reproducibility.

We set the learning-rate clipping threshold $\alpha_{max}$ = 5.0 and the gradient-scaling upper bound $\sigma_{max}$ = 5.0 in all experiments, which prevents the inverse-active-ratio scaling in Eq. (5) and the group-level gradient coefficient in Eq. (6) from producing excessively large updates and thus stabilizes training.

We compare four fine-tuning methods under this unified setting. Full SFT updates all model parameters using DeepSpeed ZeRO Stage 3 with a weight decay of 0.1. ESFT selects task-relevant experts via a gate score threshold of 0.2 and fine-tunes only those experts, using DeepSpeed ZeRO Stage 1 with a weight decay of 0.1. LoRA applies low-rank adaptation with rank 32, $\alpha$ = 64, and dropout 0.05, targeting the MLP layers (gate\_proj, up\_proj, down\_proj), trained with DeepSpeed ZeRO Stage 1 and a weight decay of 0.1. NSFT performs fine-grained sub-expert updates with dynamic gradient scaling guided by pre-computed gate energy distributions. NSFT uses DeepSpeed ZeRO Stage 1 with weight decay set to 0 to prevent frozen parameters from drifting, and employs MoE-aware learning rate scaling with an EMA decay factor of 0.9 and a scale update interval of 5 steps. By default, NSFT uses a group size of 16 and a gate score threshold of 0.2; we also ablate on group sizes of {1, 2, 4, 8, 32} and thresholds of {0.2, 0.3, 0.4}.

\subsection{Datasets}

\subsubsection{Dataset Details} 
We evaluate our method across seven diverse and challenging downstream tasks spanning distinct domains: 

\begin{itemize}
    \item TableQA (TableInstruct), a table understanding benchmark requiring structured reasoning over tabular data;
    \item SciRIFF, a scientific instruction-following dataset demanding domain-specific reasoning across scientific literature;
    \item Math (GSM8k and MATH500), evaluated on two complementary mathematical reasoning benchmarks and reported as their average score;
    \item Code (HumanEval and MBPP), evaluated on two widely-used code generation benchmarks and reported as their average score;
    \item RAG (PubMedQA and Health\_Claims), a retrieval-augmented QA task evaluated on two benchmarks and reported as their average, requiring the model to synthesize answers from long retrieved contexts in the biomedical domain.
\end{itemize}

All datasets are formatted in the standard chat-based message format and pre-tokenized offline with a maximum sequence length of 8,192 tokens. We adopt a bin-packing strategy to concatenate multiple samples into packed sequences for efficient training.

\subsubsection{On the choice of evaluation datasets and benchmarks}
Our benchmark suite is deliberately chosen to be complex and challenging — spanning structured reasoning (TableQA), domain-specific scientific and medical knowledge (SciRIFF, MMedC), long-context retrieval comprehension (RAGQA), multi-step mathematical and code reasoning (Math, Code), and cross-lingual generalization (Multilingual). These tasks demand deep adaptation of model knowledge and are far more likely to expose the limitations of parameter-efficient approaches when compared to full fine-tuning. By evaluating under such demanding conditions, we aim to provide a more rigorous and convincing assessment of our proposed NSFT method, demonstrating that it can maintain competitive performance even in scenarios where conventional PEFT methods typically fall short.

\section{Additional Results on TableQA}

\label{tableqa_appendix}

\begin{table*}[t]
\centering
\scriptsize
\setlength{\tabcolsep}{2.5pt}
\renewcommand{\arraystretch}{1.18}
\begin{adjustbox}{width=\textwidth}
\begin{tabular}{llccccccc}
\toprule
\multirow{2}{*}{\textbf{Method}} & \multirow{2}{*}{\textbf{\shortstack{Train. \\ \#Params}}} 
& \multicolumn{4}{c}{\textbf{TableQA Subtasks}} 
& \multirow{2}{*}{\textbf{\shortstack{PoT \\ Avg.}}} 
& \multirow{2}{*}{\textbf{\shortstack{General \\ Avg.}}} \\
\cmidrule(lr){3-6}
& & \rotatebox{45}{\textbf{Instruction Type}} & \rotatebox{45}{\textbf{Fact Checking}} & \rotatebox{45}{\textbf{Numerical Reasoning}} & \rotatebox{45}{\textbf{Data Analysis \& Vis.}} & & \\
\midrule
Ling-mini-2.0 (Base)  & \cellcolor{traincol} --       & 32.29 & 29.72 &  7.65 & 12.00 & 19.79 & 66.45 \\
Full FT                 & \cellcolor{traincol} 100\%    & 48.96 & 53.40 & 26.07 & 30.00 & \textbf{39.34} & 62.59 \\
\specialrule{0.4pt}{1.5pt}{2pt}
LoRA-r32 MLP           & \cellcolor{traincol} 7.38\%   &  9.38 &  8.56 &  3.99 & 16.00 &  6.41 & 63.63 \\
LoRA-r32 MLP+QKVO      & \cellcolor{traincol} 7.45\%   & 39.58 & 40.30 & 16.39 & 20.00 & 28.71 & 64.09 \\
LoRA-r64 MLP           & \cellcolor{traincol} 14.77\%  & 11.46 & 12.09 &  7.45 & 18.00 &  9.55 & 63.38 \\
LoRA-r64 MLP+QKVO      & \cellcolor{traincol} 14.84\%  & 43.75 & 43.83 & 19.64 & 16.00 & 32.00 & 63.90 \\
LoRA-r128 MLP          & \cellcolor{traincol} 29.54\%  & 20.83 & 24.18 & 13.75 & 24.00 & 18.43 & 62.89 \\
LoRA-r128 QKVO         & \cellcolor{traincol} 0.15\%   & 27.08 & 15.87 &  7.72 & 18.00 & 13.04 & 61.06 \\
LoRA-r128 MLP+QKVO     & \cellcolor{traincol} 29.69\%  & 48.96 & 48.61 & 21.30 & 26.00 & 35.35 & 63.09 \\
\specialrule{0.4pt}{1.5pt}{2pt}
ESFT-th0.2            & \cellcolor{traincol} 8.92\%   & 37.50 & 21.91 &  9.76 & 24.00 & 17.68 & 67.03 \\
ESFT-th0.3            & \cellcolor{traincol} 5.75\%   & 34.38 & 32.75 &  8.37 & 12.00 & 21.65 & 67.08 \\
ESFT-th0.4            & \cellcolor{traincol} 3.35\%   & 25.00 & 24.18 &  7.12 & 16.00 & 16.31 & 66.29 \\
\specialrule{0.4pt}{1.5pt}{2pt}
NSFT-G16-th0.2        & \cellcolor{traincol} 3.05\%   & 42.71 & 39.40 & 11.53 & 26.59 & 26.59 & 66.53 \\
NSFT-G16-th0.3        & \cellcolor{traincol} 5.45\%   & 30.21 & 39.29 & 16.48 & 27.27 & 27.27 & 65.99 \\
NSFT-G16-th0.4        & \cellcolor{traincol} 8.57\%   & 33.33 & 45.09 & 12.18 & 15.00 & 28.54 & 66.31 \\
NSFT-G32-th0.2        & \cellcolor{traincol} 3.09\%   & 30.21 & 39.55 &  8.03 & 18.00 & 24.11 & 66.40 \\
NSFT-G8-th0.2         & \cellcolor{traincol} 2.98\%   & 57.29 & 46.85 & 16.95 & 18.00 & 33.77 & 66.04 \\
NSFT-G4-th0.2         & \cellcolor{traincol} 2.86\%   & 48.96 & 50.88 & 16.34 & 18.00 & 34.44 & 66.18 \\
NSFT-G2-th0.2         & \cellcolor{traincol} 2.66\%   & 43.75 & 45.59 & 18.36 & 16.00 & 32.28 & 66.04 \\
NSFT-G1-th0.2         & \cellcolor{traincol} 2.40\%   & 43.75 & 41.31 & 16.98 & 16.00 & 29.83 & 64.98 \\
\bottomrule
\end{tabular}
\end{adjustbox}
\caption{Detailed TableQA results on Ling-mini-2.0. We compare Full FT, LoRA with different ranks and target modules, ESFT under different selection thresholds, and NSFT under different thresholds and group sizes. The table reports performance on four TableQA subtasks, the PoT average, and the average performance on general benchmarks. Bold numbers indicate the best result within each column.}
\label{tab:tableqa_appendix}
\end{table*}

Table~\ref{tab:tableqa_appendix} provides a more detailed comparison on TableQA using Ling-mini-2.0. We include LoRA variants with different ranks and target modules, ESFT under different selection thresholds, and NSFT under different thresholds and group sizes. These results further validate the advantage of fine-grained sub-expert adaptation.

\paragraph{NSFT achieves the strongest efficiency--performance trade-off on TableQA.}
Compared with LoRA and ESFT, NSFT obtains competitive or stronger TableQA performance while using substantially fewer trainable parameters. In particular, NSFT-G4-th0.2 reaches a PoT average of 34.44 with only 2.86\% trainable parameters, outperforming all ESFT variants and approaching the strongest LoRA variant, LoRA-128 MLP+QKVO, which obtains a PoT average of 35.35 but requires 29.69\% trainable parameters. This indicates that updating fine-grained sub-experts can recover most of the task-specific adaptation ability of much larger PEFT configurations with nearly an order of magnitude fewer trainable parameters.

\paragraph{NSFT is consistently stronger than expert-level sparse tuning.}
Across different thresholds, NSFT clearly outperforms ESFT on TableQA. The best ESFT variant reaches a PoT average of 21.65, whereas multiple NSFT variants exceed this score by a large margin. For example, NSFT-G16-th0.2, NSFT-G8-th0.2, NSFT-G4-th0.2, NSFT-G2-th0.2, and NSFT-G1-th0.2 all outperform ESFT-th0.3. This confirms that expert-level selection is too coarse for TableQA: updating whole experts still includes many task-irrelevant internal channels, while sub-expert selection can allocate the update budget to more relevant parts inside experts.

\paragraph{NSFT better preserves general capability than full fine-tuning and large LoRA variants.}
Although Full FT obtains the highest PoT average, it significantly reduces the general average from 66.45 to 62.59. Large LoRA variants also tend to underperform the base model on general benchmarks. In contrast, NSFT maintains competitive general capability, with several variants achieving general averages around 66.0--66.5 while substantially improving TableQA performance. This shows that fine-grained sub-expert tuning provides a better balance between domain specialization and general capability retention.

\paragraph{Discussion on ESFT degradation.}
Table~\ref{tab:tableqa_appendix} shows that ESFT underperforms the Ling-mini-2.0 base model on TableQA, while LoRA variants that only update MoE-related modules also provide limited gains without a much larger trainable budget. A likely reason is that Ling-mini itself adopts a finer expert partitioning than OLMoE, using 256 experts instead of 64. This architectural design indicates that Ling-mini distributes its computation across more fine-grained expert units. Accordingly, adaptation should also be performed at a finer granularity: updating entire selected experts may still be too coarse, while sub-expert-level updates can better match the model's internal organization.

This highlights the need to align adaptation granularity with model granularity. NSFT further decomposes experts into structured sub-expert groups and updates only the most responsive ones, providing a finer update unit that better matches Ling-mini's architecture. This explains why NSFT achieves stronger TableQA performance with fewer trainable parameters and suggests that fine-grained MoE backbones may benefit more from sub-expert-level adaptation.

\section{Robustness across Selection Thresholds}
\label{threshold_appendix}
\begin{table}[t]
\centering
\small
\setlength{\tabcolsep}{6pt}
\renewcommand{\arraystretch}{1.12}
\begin{adjustbox}{width=0.8\textwidth}
\begin{tabular}{lccccc}
\toprule
Method & $\tau=0.1$ & $\tau=0.2$ & $\tau=0.3$ & $\tau=0.4$ & $\tau=0.5$ \\
\midrule
ESFT-OLMoE & 42.90 & 44.31 & 46.16 & 47.34 & 48.24 \\
NSFT-OLMoE & \textbf{43.32} & \textbf{45.46} & \textbf{46.81} & \textbf{47.76} & \textbf{48.78} \\
\bottomrule
\end{tabular}
\end{adjustbox}
\vspace{4pt}
\caption{Performance comparison under different selection thresholds on MMedC. We compare ESFT and NSFT variants under different selection thresholds $\tau$. Best results under each threshold are bolded. The NSFT results correspond to NSFT-SG, which uses learning-rate scaling and static gradient scaling.}
\label{tab:threshold_mmedc}
\end{table}

\begin{table}[t]
\centering
\small
\setlength{\tabcolsep}{5pt}
\renewcommand{\arraystretch}{1.12}
\begin{tabular}{lccccc}
\toprule
Method & $\tau=0.1$ & $\tau=0.2$ & $\tau=0.3$ & $\tau=0.4$ & $\tau=0.5$ \\
\midrule
ESFT-OLMoE
& 36.44 / 35.62 
& 40.15 / 38.08 
& 42.67 / 40.21 
& 42.55 / 39.59 
& 44.13 / 41.13 \\
NSFT-OLMoE
& \textbf{37.84} / \textbf{36.97}
& \textbf{43.27} / \textbf{39.99}
& \textbf{43.67} / \textbf{40.74}
& \textbf{44.36} / \textbf{41.06}
& \textbf{44.67} / \textbf{41.40} \\
\bottomrule
\end{tabular}
\vspace{4pt}
\caption{Performance comparison under different selection thresholds on SciRiFF under 4k / 8k settings. Each cell reports results under 4k and 8k training lengths, respectively. Best results under each threshold and setting are bolded. The NSFT results correspond to NSFT-SG, which uses learning-rate scaling and static gradient scaling.}
\label{tab:threshold_sciriff}
\end{table}

\paragraph{Discussion.}
Tables~\ref{tab:threshold_mmedc} and~\ref{tab:threshold_sciriff} report detailed results under different selection thresholds. NSFT consistently outperforms ESFT across both MMedC and SciRiFF, showing that sub-expert-level selection is more robust than expert-level sparse tuning under different sparsity budgets. Performance generally improves as $\tau$ increases, since more task-relevant units are retained. These results suggest that the optimal granularity depends on the target domain, but the advantage of fine-grained sub-expert selection remains consistent.

\section{Group Size Sensitivity across MoE Backbones}
\label{group_appendix}

\begin{table}[t]
\centering
\small
\setlength{\tabcolsep}{7pt}
\renewcommand{\arraystretch}{1.2}
\begin{adjustbox}{width=\textwidth}
\begin{tabular}{lccccccc}
\toprule
\textbf{Group Size} & \textbf{1} & \textbf{2} & \textbf{4} & \textbf{8} & \textbf{16} & \textbf{32} &  \textbf{1024}\\
\midrule
Number of involved experts $a$ & \textbf{1024} & 730 & 526 & 376 & 288 & 230 & 173 \\
Average activated ratio within selected experts $b$ & 12.09\% & 18.85\% & 28.09\% & 40.91\% & 54.69\% & 69.46\% & \textbf{100.0\%} \\
TableQA PoT & 29.83 & 32.28 & \textbf{34.44} & 33.77 & 26.59 & 24.11 & 17.68 \\
\bottomrule
\end{tabular}
\end{adjustbox}
\vspace{4pt}
\caption{Relationship between group size, activated expert coverage, and TableQA PoT performance on Ling-mini-2.0.}
\label{tab:group_activation_tableqa}
\end{table}

\paragraph{The optimal group size is backbone-dependent.}
Although our main experiments on OLMoE suggest that a moderate group size such as $G=16$ provides a strong balance between flexibility and stability, the results on Ling-mini-2.0 show that the optimal group size can shift across MoE backbones. As shown in Table~\ref{tab:group_activation_tableqa}, when applying NSFT to TableQA on Ling-mini-2.0, the best PoT performance is achieved at $G=4$, reaching 34.44. This differs from the trend observed on OLMoE, where larger groups such as $G=16$ are often more favorable. This indicates that group size should not be treated as a universal constant, but rather as a backbone- and task-dependent hyperparameter.

\paragraph{Smaller groups provide broader expert coverage on Ling-mini-2.0.}
Table~\ref{tab:group_activation_tableqa} reveals a clear relationship between group size, expert coverage, and task performance. As $G$ decreases from 32 to 1, the number of involved experts increases substantially, from 230 to 1024. This suggests that smaller groups allow NSFT to select sub-experts from a much broader set of experts, thereby covering more diverse computation paths. For Ling-mini-2.0 on TableQA, this broader coverage appears to be important, likely because TableQA requires heterogeneous capabilities such as instruction understanding, fact checking, numerical reasoning, and data analysis. Therefore, a smaller group size can better capture the distributed expert patterns needed by this task.

\paragraph{Connection to expert-level tuning.}
It is also worth noting that when the group size becomes equal to the expert intermediate dimension, i.e., $G=1024$ in Ling-mini-2.0, each expert contains only one group. In this extreme case, sub-expert selection degenerates into expert-level selection, which is essentially equivalent to ESFT. As shown in Table~\ref{tab:group_activation_tableqa}, this setting involves 173 experts and achieves a TableQA PoT score of 17.68, much lower than the best sub-expert setting with $G=4$. This result further confirms that expert-level granularity is too coarse: updating an entire expert introduces substantial redundant parameters, while decomposing experts into finer-grained sub-experts enables more precise and effective adaptation.

These results provide a practical guideline for choosing the group size. When task-relevant computation is distributed across many experts, smaller groups such as $G=4$ or $G=8$ offer broader expert coverage and greater selection flexibility. When useful activations are concentrated within fewer experts, moderately larger groups such as $G=16$ can be more stable and efficient. We therefore treat $G$ as a lightweight hyperparameter and select it with a small validation sweep, using expert coverage and intra-expert activation ratio as diagnostic signals to avoid both overly fragmented and overly coarse updates.

\section{Additional Analysis for Intra-expert Sparsity in ESFT}
\label{intra_expert_appendix}

\begin{figure*}[t!]
  \centering
  \includegraphics[width=\linewidth]{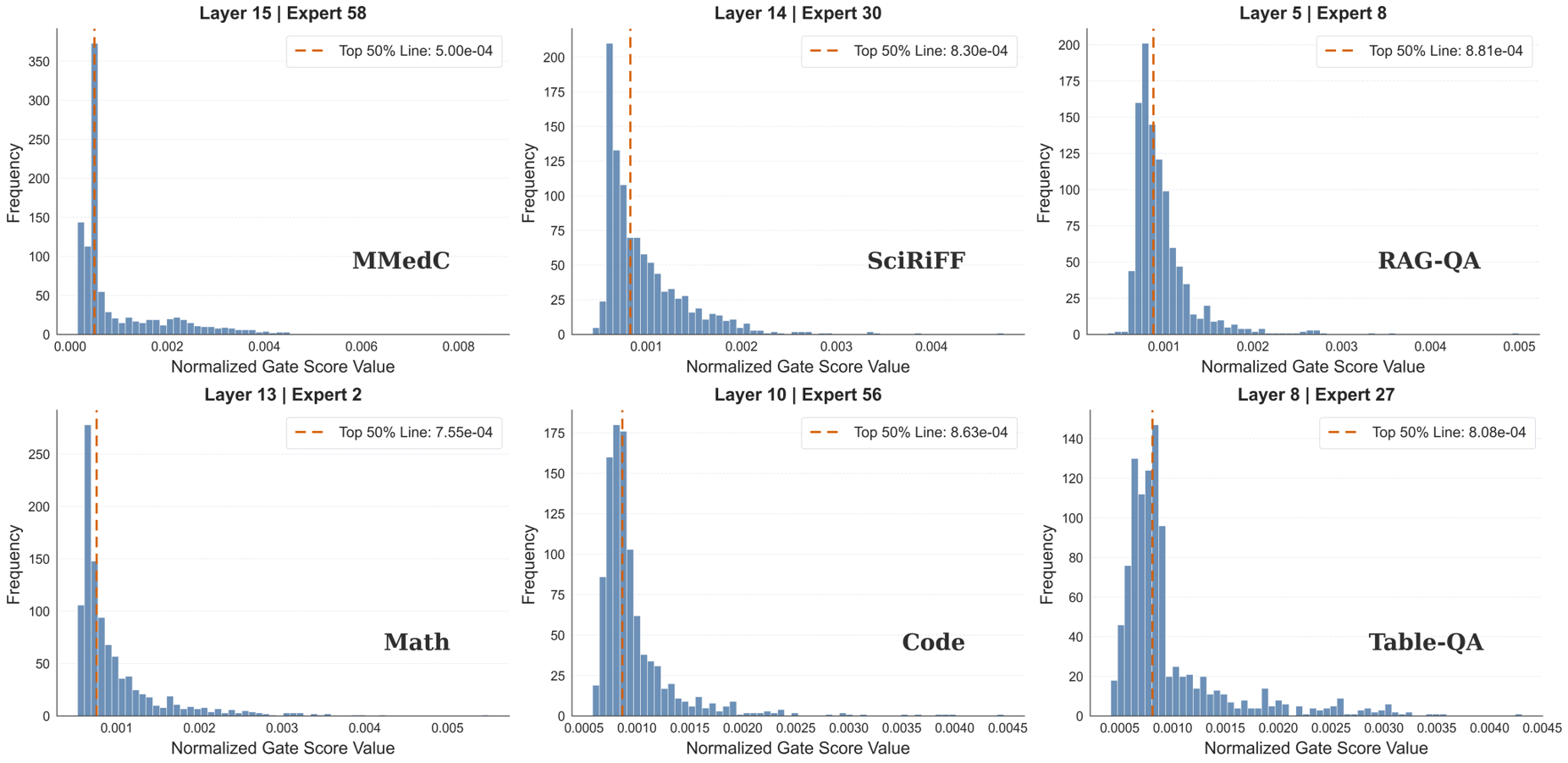}
  \caption{Intra-expert activation sparsity across domains. Each subplot shows the normalized channel-wise gate-score distribution of a representative OLMoE expert. The dashed orange line marks the top-50\% cumulative activation threshold. The strong concentration near zero across domains indicates that only a small subset of channels is highly activated inside selected experts.}
  \label{fig:gini_across_layers}
  \vspace{-0.2cm}
\end{figure*}

\begin{figure*}[t!]
  \centering
  \includegraphics[width=\linewidth]{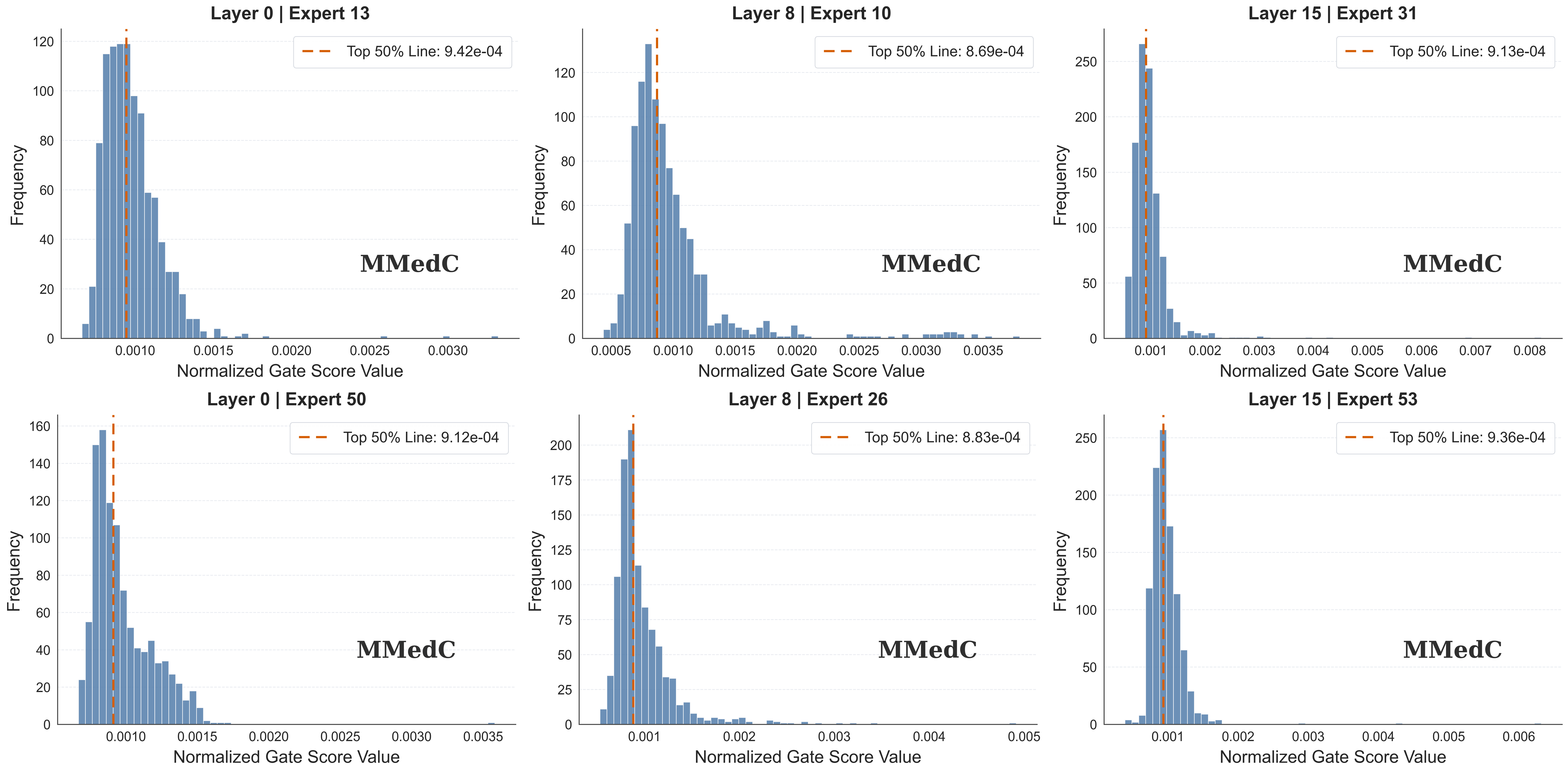}
  \caption{Channel-wise gate-score distributions for randomly sampled experts from shallow, middle, and deep layers on MMedC, with two experts sampled from each depth range. The dashed orange line marks the threshold covering the top 50\% cumulative activation mass.}
  \label{fig:expert_histograms}
  \vspace{-0.2cm}
\end{figure*}

\subsection{Intra-expert activation sparsity across domains}
Figure~\ref{fig:gini_across_layers} visualizes the channel-wise gate-score distributions of representative OLMoE experts from different layers and domains. Across all datasets, the distributions are highly concentrated near zero, while only a small fraction of channels receive relatively large activation scores. The dashed orange line marks the threshold covering the top 50\% cumulative activation mass, and it consistently lies in a low-value region. This indicates that even within activated or selected experts, the effective activation mass is dominated by a small subset of internal channels. Such a pattern is observed across diverse domains, including medical, scientific, retrieval, mathematical, code, and table reasoning tasks, suggesting that intra-expert sparsity is a general property rather than a domain-specific artifact. These findings further support our motivation for sub-expert selection: updating an entire selected expert, as in expert-level sparse tuning, inevitably includes many weakly activated channels, whereas fine-grained sub-expert selection can focus the trainable budget on the most responsive internal components.

\subsection{Consistent sparsity across experts}
To further verify that intra-expert sparsity is not limited to a few special cases, Figure~\ref{fig:expert_histograms} visualizes the distributions of normalized gate scores for randomly sampled experts from shallow, middle, and deep layers. Specifically, we randomly select two experts from each of the shallow, middle, and deep layers. Despite differences in layer depth and expert identity, all sampled experts exhibit a highly similar pattern: most channels are concentrated in the near-zero region, while only a small fraction of channels form a long tail with relatively larger activations. The dashed line marking the threshold that covers the top 50\% cumulative activation mass also lies in a very small-value region across all cases. 

These results confirm that strong intra-expert sparsity is a stable and widespread phenomenon, rather than an artifact of a particular layer or expert. This consistency further motivates our sub-expert selection strategy, which allocates updates to the most responsive internal channel groups instead of updating the full expert uniformly.

\section{Additional Results on Static and Dynamic Gradient Scaling}
\label{static_dynamic_appendix}

In the main text, we report the averaged results over three datasets to summarize the overall effect of static and dynamic gradient scaling. Here, we provide detailed results on each dataset for a more fine-grained comparison.

Tables~\ref{tab:medical_scaling_ablation} and~\ref{tab:sciriff_scaling_ablation} provide additional comparisons between static gradient scaling and dynamic gradient scaling on MMedC and SciRiFF. Overall, dynamic scaling generally improves in-domain adaptation over the static variant, but the best update frequency varies across tasks. On MMedC, updating the scaling mask every training step achieves the best weighted average and general-benchmark average, while updating every five steps gives the best unweighted average. On SciRiFF, the five-step update performs better, achieving the highest weighted and unweighted validation averages under both 4096 and 8192 settings. These results suggest that dynamic scaling can better track the evolving importance of selected sub-experts, while a less frequent update schedule can be beneficial when online statistics are noisy.

\begin{table*}[t]
\centering
\scriptsize
\setlength{\tabcolsep}{2.pt}
\renewcommand{\arraystretch}{1.12}
\begin{adjustbox}{width=\textwidth}
\begin{tabular}{lccccccccc}
\toprule
\textbf{Method} &  \textbf{Chinese} & \textbf{English} & \textbf{French} & \textbf{Japanese} & \textbf{Spanish} & \textbf{Russian} & \textbf{W-AVG} & \textbf{UW-AVG} & \textbf{General Avg.} \\
\midrule
NSFT-SG & 46.82 & 49.73 & 41.80 & 25.13 & 42.74 & 59.77 & 45.46 & 44.33 & 45.43 \\
NSFT-DSG\_iter1 & 47.23 & 50.04 & \textbf{45.34} & 23.62 & \textbf{42.78} & 59.38 & \textbf{45.89} & 44.73 & \textbf{45.75} \\
NSFT-DSG\_iter5 & \textbf{48.25} & \textbf{48.23} & 43.73 & \textbf{26.63} & 42.16 & \textbf{60.16} & 45.81 & \textbf{44.86} & 45.51 \\
\bottomrule
\end{tabular}
\end{adjustbox}
\caption{Static vs. dynamic gradient scaling on MMedC multilingual subtasks using OLMoE. \textbf{DSG} denotes dynamic gradient scaling, and \textbf{DSG-EMA} applies EMA smoothing. The table reports language-wise results, weighted/unweighted averages, and general-benchmark average.}
\label{tab:medical_scaling_ablation}
\end{table*}

\begin{table*}[t]
\centering
\scriptsize
\setlength{\tabcolsep}{2pt}
\renewcommand{\arraystretch}{1.12}
\begin{adjustbox}{width=\textwidth}
\begin{tabular}{lccccccccccccc}
\toprule
\multirow{2}{*}{\textbf{Method}} 
& \multicolumn{2}{c}{\textbf{BioASQ} (64)} 
& \multicolumn{2}{c}{\textbf{BioRED} (100)} 
& \multicolumn{2}{c}{\textbf{Evidence\_Inference} (250)} 
& \multicolumn{2}{c}{\textbf{SciERC} (50)} 
& \multicolumn{2}{c}{\textbf{W-AVG}} 
& \multicolumn{2}{c}{\textbf{UW-AVG}} 
& \multirow{2}{*}{\textbf{\shortstack{General \\ Avg.}}} \\
\cmidrule(lr){2-3} \cmidrule(lr){4-5} \cmidrule(lr){6-7} \cmidrule(lr){8-9} \cmidrule(lr){10-11} \cmidrule(lr){12-13}
& 4096 & 8192 & 4096 & 8192 & 4096 & 8192 & 4096 & 8192 & 4096 & 8192 & 4096 & 8192 & \\
\midrule
NSFT-SG & 62.17 & 43.41 & 72.73 & 71.88 & 26.63 & 25.61 & 43.35 & 43.75 & 43.27 & 39.99 & 51.22 & 46.16 & \textbf{45.55} \\
NSFT-DSG\_iter1 & 63.06 & 42.73 & 72.43 & 72.21 & 25.36 & 25.25 & \textbf{43.86} & \textbf{44.22} & 42.70 & 39.83 & 51.18 & 46.10 & 45.05 \\
NSFT-DSG\_iter5 & \textbf{63.52} & \textbf{44.55} & \textbf{73.50} & \textbf{73.54} & \textbf{28.46} & 28.68 & 41.23 & 42.70 & \textbf{44.38} & \textbf{42.05} & \textbf{51.68} & \textbf{47.37} & 45.46 \\
\bottomrule
\end{tabular}
\end{adjustbox}
\caption{Static vs. dynamic gradient scaling on SciRiFF subtasks using OLMoE.}
\label{tab:sciriff_scaling_ablation}
\end{table*}

\section{Low-Budget and Budget-Matched Comparisons}
\label{app:low_budget_comparison}

\paragraph{Lower-budget regime.}
Table~\ref{tab:low_budget_frontier} extends the comparison between ESFT and
NSFT to selection thresholds below the default $\tau=0.2$. At
$\tau=0.05$, the coarse selection unit of ESFT causes nearly every layer to
retain only one complete expert, which represents a practical lower bound on
its parameter budget. In contrast, NSFT can continue reducing the budget by
selecting groups within experts. NSFT consistently obtains higher MMedC
performance while updating fewer parameters across all three thresholds,
demonstrating a stronger accuracy--parameter trade-off in the low-budget
regime.

\begin{table}[!htbp]
    \centering
    \small
    \setlength{\tabcolsep}{5pt}
    \begin{tabular}{c cc cc c}
        \toprule
        & \multicolumn{2}{c}{ESFT} & \multicolumn{2}{c}{NSFT}
        & $\Delta$ \\
        \cmidrule(lr){2-3}\cmidrule(lr){4-5}
        Threshold & Train. & MMedC Avg. & Train. & MMedC Avg.
        & NSFT$-$ESFT \\
        \midrule
        $\tau=0.20$ & 5.64\% & 44.31 & 4.87\% & \textbf{45.74} & +1.43 \\
        $\tau=0.10$ & 3.09\% & 42.90 & 2.08\% & \textbf{43.32} & +0.42 \\
        $\tau=0.05$ & 1.55\% & 41.35 & 0.97\% & \textbf{42.30} & +0.95 \\
        \bottomrule
    \end{tabular}
    \caption{Low-budget comparison between ESFT and NSFT on MMedC. Train.
    denotes the percentage of trainable parameters.}
    \label{tab:low_budget_frontier}
\end{table}

\paragraph{Closely matched parameter budgets.}
Because ESFT selects complete experts whereas NSFT selects groups within
experts, their trainable-parameter ratios cannot generally be matched exactly
using the same threshold. We therefore adjust the NSFT threshold to
$\tau=0.22$ to obtain budgets close to those of ESFT at $\tau=0.20$.
As shown in Table~\ref{tab:budget_matched_comparison}, NSFT remains stronger on
both SciRIFF and MMedC under closely matched budgets. This result rules out a
larger trainable-parameter budget as the source of NSFT's improvement.

\begin{table}[!htbp]
    \centering
    \small
    \setlength{\tabcolsep}{4.5pt}
    \begin{tabular}{l c cc cc}
        \toprule
        & & \multicolumn{2}{c}{SciRIFF} & \multicolumn{2}{c}{MMedC} \\
        \cmidrule(lr){3-4}\cmidrule(lr){5-6}
        Method & Threshold & Train. & 4k / 8k Avg. & Train. & Avg. \\
        \midrule
        ESFT & $\tau=0.20$ & 7.46\% & 40.15 / 38.08 & 5.64\% & 44.31 \\
        NSFT & $\tau=0.20$ & 6.57\% & 40.48 / 38.88 & 4.87\% & 44.93 \\
        NSFT & $\tau=0.22$ & 7.55\% & \textbf{41.88 / 39.63}
             & 5.54\% & \textbf{45.50} \\
        \bottomrule
    \end{tabular}
    \caption{Comparison under closely matched trainable-parameter budgets.
    The NSFT threshold is adjusted to match the ESFT budget as closely as
    permitted by their different selection granularities.}
    \label{tab:budget_matched_comparison}
\end{table}

\paragraph{Weight-decay control.}
The main experiments follow the recommended settings of each baseline: ESFT
uses weight decay $0.1$, whereas NSFT uses weight decay $0$ to avoid
decay-induced updates to masked parameters. As a control, setting the ESFT
weight decay to $0$ yields an MMedC average of 44.01, compared with 44.31 under
its original weight decay of $0.1$. The result is slightly lower rather than
higher, indicating that the advantage of NSFT is not caused by the different
weight-decay settings.


\section{Alternative Sub-Expert Grouping Strategies}
\label{app:alternative_grouping}

Our default method forms structured groups from consecutive channels. This
choice does not assume that individual channels have explicit
human-interpretable semantics; rather, it provides a simple and low-cost way to
construct stable update units in the learned parameter layout. To test whether
the result depends on this choice, we compare it with three alternatives while
keeping the remaining training settings fixed: (1) random grouping, which
randomly assigns channels to groups of size 16; (2) activation-clustering
grouping, which clusters channels according to the cosine similarity of their
normalized up-projection responses over tokens; and (3) individual-channel
selection ($G=1$), which performs top-$K$ non-contiguous selection without
structured grouping.

Table~\ref{tab:grouping_strategy_ablation} shows that random and
activation-clustering groups mitigate the general-capability degradation of
individual-channel selection, but remain weaker than consecutive grouping.
The default strategy therefore provides the best overall balance between
in-domain adaptation, general capability, simplicity, and grouping overhead.

\begin{table}[!htbp]
    \centering
    \small
    \begin{tabular}{l cc}
        \toprule
        Grouping Strategy & MMedC W-Avg. & General Avg. \\
        \midrule
        Random grouping ($G=16$) & 44.91 & 43.71 \\
        Activation-clustering grouping ($G=16$) & 44.04 & 43.77 \\
        Individual channels ($G=1$) & 44.65 & 38.84 \\
        Consecutive grouping ($G=16$, default) & \textbf{44.93} & \textbf{44.41} \\
        \bottomrule
    \end{tabular}
    \caption{Comparison of alternative sub-expert grouping strategies on
    MMedC. W-Avg. denotes the weighted average over the multilingual subtasks.}
    \label{tab:grouping_strategy_ablation}
\end{table}


\section{Contributions of Selection and Adaptive Optimization}
\label{app:component_contributions}

To isolate the effect of fine-grained selection, we begin with
NSFT-G16-Select, which uses only sub-expert selection at $\tau=0.2$ and removes
all adaptive optimization strategies. We then progressively add static and
dynamic gradient scaling. The static variant includes learning-rate scaling
and a fixed gradient-scaling mask, whereas the dynamic variant periodically
refreshes the mask using updated activation statistics.

As shown in Table~\ref{tab:component_ablation}, selection alone already
outperforms ESFT on both MMedC and SciRIFF-8k, confirming that the primary gain
comes from refining the update unit from experts to sub-experts. Static and
dynamic scaling then provide further, complementary improvements, with dynamic
scaling giving the largest gain on SciRIFF-8k.

\begin{table}[!htbp]
    \centering
    \small
    \begin{tabular}{l cc}
        \toprule
        Method & MMedC Avg. & SciRIFF-8k Avg. \\
        \midrule
        ESFT & 44.31 & 38.08 \\
        NSFT-G16-Select & 44.93 & 38.74 \\
        \quad + Static Gradient Scaling & 45.46 & 39.99 \\
        \quad + Dynamic Gradient Scaling & \textbf{45.81} & \textbf{42.05} \\
        \bottomrule
    \end{tabular}
    \caption{Step-by-step ablation separating sub-expert selection from
    adaptive optimization.}
    \label{tab:component_ablation}
\end{table}

\section{Effect of the Entropy-Adaptive Exponent}
\label{app:gamma_ablation}

The entropy-adaptive exponent $\gamma$ smooths group-wise gradient scaling:
when group energies are relatively uniform, $\gamma$ approaches 1 and
preserves their contrast; when the distribution is concentrated, $\gamma$
moves toward 0.5 and suppresses extreme scaling coefficients. We evaluate its
effect using the $G=32$ model with static gradient scaling on MMedC.
Table~\ref{tab:gamma_ablation} shows that replacing the adaptive exponent with
a fixed exponent of 1.0 reduces the MMedC average from 45.06 to 44.87. Although
the gain is modest, it consistently supports adaptive smoothing over the raw
energy-ratio scaling rule.

\begin{table}[!htbp]
    \centering
    \small
    \begin{tabular}{l cc}
        \toprule
        Method & Exponent in Gradient Scaling & MMedC Avg. \\
        \midrule
        NSFT w/o adaptive $\gamma$ & $1.0$ & 44.87 \\
        NSFT w/ adaptive $\gamma$ & Entropy-adaptive & \textbf{45.06} \\
        \bottomrule
    \end{tabular}
    \caption{Ablation of the entropy-adaptive exponent using $G=32$ and
    static gradient scaling on MMedC.}
    \label{tab:gamma_ablation}
\end{table}

\section{Training Efficiency and Dynamic-Scaling Overhead}
\label{app:training_efficiency}

Table~\ref{tab:training_cost} reports wall-clock training time, peak GPU
memory, and MMedC performance. Wall-clock time is measured using eight NVIDIA
A100 GPUs, while peak per-GPU memory is measured using four A100 GPUs under
the same MMedC setting. NSFT-selection has a training time close to ESFT and
already achieves higher accuracy, showing that sub-expert selection itself
introduces little additional time overhead. All NSFT variants also retain
memory consumption close to ESFT and substantially below Full FT.

The main additional cost of full NSFT comes from collecting updated activation
statistics and refreshing the scaling mask. Removing dynamic updates reduces
training time from 35m 29s to 20m 38s without increasing memory usage. Dynamic
scaling should therefore be viewed as an optional performance enhancement:
users may choose selection-only, static-scaling, or full dynamic-scaling NSFT
according to their compute budget. More frequent updates track rapidly changing
sub-expert importance more closely, whereas less frequent updates can be more
stable when the activation statistics are noisy or heterogeneous.

\begin{table}[!htbp]
    \centering
    \small
    \setlength{\tabcolsep}{6pt}
    \begin{tabular}{l c c c}
        \toprule
        Method & MMedC Avg. & Time (8$\times$A100) & Peak Memory (4$\times$A100) \\
        \midrule
        Full FT & \textbf{52.12} & 25m 24s & 67 GB \\
        ESFT & 44.31 & 17m 19s & 24 GB \\
        NSFT-selection & 44.93 & 18m 30s & 26 GB \\
        NSFT-static & 45.46 & 20m 38s & 26 GB \\
        NSFT & 45.81 & 35m 29s & 26 GB \\
        \bottomrule
    \end{tabular}
    \caption{Training cost and performance on MMedC. Wall-clock time and peak
    per-GPU memory are measured with eight and four A100 GPUs, respectively.}
    \label{tab:training_cost}
\end{table}

\section{Limitations}
\label{limitation}
Although NSFT achieves strong results across representative MoE backbones and diverse downstream tasks, several aspects deserve further exploration. First, we evaluate NSFT on two competitive MoE models, OLMoE-7B and Ling-mini-2.0-16B, which already cover different scales and architectural characteristics. The consistent improvements across these models suggest that sub-expert-level adaptation is broadly effective. Future work may further examine NSFT on larger and more heterogeneous MoE LLMs to better characterize its scaling behavior.

Second, NSFT uses the group size $G$ and selection threshold $\tau$ to control the granularity and budget of sub-expert selection. Our experiments show stable gains across multiple choices of these hyperparameters, and the default setting works well in practice. Still, for new domains or substantially different MoE architectures, a lightweight validation sweep may help identify the best efficiency--performance trade-off.

\section{Broader Impacts}
\label{broader_impacts}
This work improves the parameter efficiency of adapting MoE LLMs to downstream domains, which can reduce computational cost, memory usage, and energy consumption compared with full-parameter fine-tuning. This may make domain-specific model adaptation more accessible to users with limited hardware resources.

However, efficient adaptation may also lower the barrier to customizing powerful models for harmful or misleading purposes. Moreover, domain-specific fine-tuning can amplify biases, errors, or unsafe behaviors present in the training data. We therefore recommend careful data curation, safety evaluation, and deployment monitoring, especially for high-stakes applications.

\end{document}